\documentclass[journal]{IEEEtran}

\usepackage{times}
\usepackage{helvet}
\usepackage{courier}
\usepackage{url}
\usepackage{graphicx}
\usepackage{mathrsfs}
\usepackage{subfigure}
\usepackage{booktabs}
\usepackage{amsfonts}
\usepackage{amsmath}
\usepackage{color}
\usepackage{cite}
\usepackage{amssymb}
\usepackage{multirow}

\begin{document}

\title{An Adversarial Learning Approach to Medical Image Synthesis for Lesion Detection}

\author{Liyan~Sun,
        Jiexiang~Wang,
        Yue~Huang,
        Xinghao~Ding,
        Hayit~Greenspan$^{\dagger}$,
        and~John~Paisley$^{\ddagger}$
\thanks{This work was supported in part by the National Natural Science Foundation of China under Grants 61571382, 81671766, 61571005, 81671674, U1605252, 61671309, in part by the Guangdong Natural Science Foundation under Grant 2015A030313007, in part by the Fundamental Research Funds for the Central Universities under Grant 20720160075, 20720180059, in part by the National Natural Science Foundation of Fujian Province, China under Grant 2017J01126. (Corresponding author: Xinghao Ding)}
\thanks{L. Sun, J. Wang, Y. Huang and X. Ding was with the School of Information Science and Engineering, Xiamen University, Xiamen, Fujian, 361005, China, corresponding to: dxh@xmu.edu.cn.}
\thanks{$^{\dagger}$ H. Greenspan was with Department of Biomedical Engineering, Tel Aviv University, Israel}
\thanks{$^{\ddagger}$ J. Paisley was with Department of Electrical Engineering, Columbia University, New York, NY, USA}
}

\markboth{Journal of \LaTeX\ Class Files,~Vol.~14, No.~8, August~2015}%
{Shell \MakeLowercase{\textit{et al.}}: Bare Demo of IEEEtran.cls for IEEE Journals}

\maketitle

\begin{abstract}
  The identification of lesion within medical image data is necessary for diagnosis, treatment and prognosis. Segmentation and classification approaches are mainly based on supervised learning with well-paired image-level or voxel-level labels. However, labeling the lesion in medical images is laborious requiring highly specialized knowledge. We propose a medical image synthesis model named \textit{abnormal-to-normal translation generative adversarial network} (ANT-GAN) to generate a normal-looking medical image based on its abnormal-looking counterpart without the need for paired training data. Unlike typical GANs, whose aim is to generate realistic samples with variations, our more restrictive model aims at producing a normal-looking image corresponding to one containing lesions, and thus requires a special design. Being able to provide a ``normal'' counterpart to a medical image can provide useful side information for medical imaging tasks like lesion segmentation or classification validated by our experiments. In the other aspect, the ANT-GAN model is also capable of producing highly realistic lesion-containing image corresponding to the healthy one, which shows the potential in data augmentation verified in our experiments.
\end{abstract}


\IEEEpeerreviewmaketitle

\section{Introduction}
Lesions can occur in body tissue as a result of various factors including trauma, infection or cancer. Medical imaging techniques such as magnetic resonance imaging (MRI) and computational tomography (CT) provide detailed information for diagnosing such lesions \cite{1}. With more efficient medical imaging systems being deployed beyond advanced societies, demands on radiologists have also been increasing. Automatic medical analysis systems can help lower the human expert barrier and expedite the diagnosis and treatment process \cite{2}.

However, in the current medical image analysis paradigm, machines and human experts differ in their approach. Specifically, radiologists are well-trained using many healthy and unhealthy medical images and transfer their learned internal representations to new images. Experts search for abnormal regions that differ from their prior knowledge bank of ``healthiness'' when mentally segmenting the lesions. As for machines, usually a function mapping of the unhealthy medical image to a certain label is learned in a supervised way, either at the image-level or the voxel-level. Since images containing lesions constitute only a small portion of available scans, information in the lesion-free image is often overlooked. Furthermore, the size of medical image datasets is usually limited because labeling requires specialized expertise and is laborious. Such data imbalance and scarcity impacts the performance of medical image analysis models negatively and motivates us to more sufficiently utilize the ``healthy" images
containing valuable prior information on the appearance of healthy brain structure. We seek to imitate the expert by building a knowledge base of healthy medical images to aid improving diagnostic performance. Predicting a fake healthy version of an image containing lesions can aid in automatic medical image analysis tasks such as segmentation and provide doctors with additional diagnostic information.

\begin{figure}[t]
   \subfigure[\scriptsize real tumor MRI]     {\label {figure1a} \includegraphics[height=0.145\textwidth,width=0.145\textwidth]{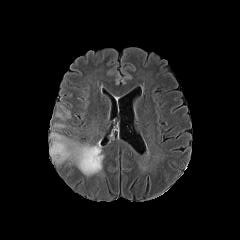}}
   \subfigure[\scriptsize generated healthy]   {\label {figure1b} \includegraphics[height=0.145\textwidth,width=0.145\textwidth]{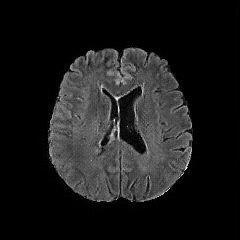}}
   \subfigure[\scriptsize difference (a)$\&$(b)]         {\label {figure1c} \includegraphics[height=0.145\textwidth,width=0.145\textwidth]{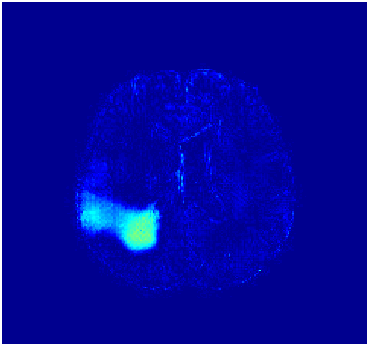}}\\
   \subfigure[\scriptsize real healthy MRI]     {\label {figure1d} \includegraphics[height=0.145\textwidth,width=0.145\textwidth]{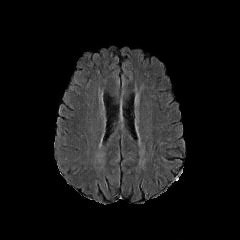}}
   \subfigure[\scriptsize generated healthy]   {\label {figure1e} \includegraphics[height=0.145\textwidth,width=0.145\textwidth]{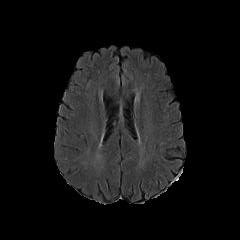}}
   \subfigure[\scriptsize difference (d)$\&$(e)]         {\label {figure1f} \includegraphics[height=0.145\textwidth,width=0.145\textwidth]{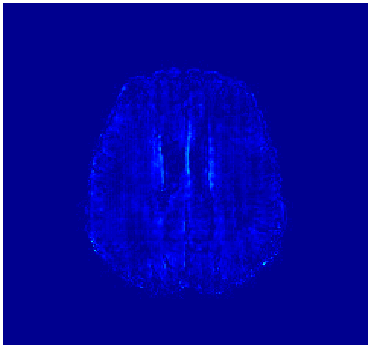}}
   \caption{Results produced by our model. Lesions are isolated, while healthy regions pass through with minimal modification.}
\label{figure1}
\end{figure}

Deep neural networks are the state-of-the-art for supervised learning in computer vision and medical imaging tasks such as image classification \cite{40,48}, segmentation \cite{3,50,51}, object detection \cite{41} and medical image reconstruction \cite{42}. However, in real clinical practice obtaining pairs of normal and abnormal images is unrealistic, since only one can exist at a time, and data augmentation methods are not available here. Thus, instead of formulating a supervised learning framework, we develop an ``abnormal-to-normal translation generative adversarial network'' (ANT-GAN) model to predict what a lesion-free image should look like that corresponds to an input image; if the model doesn't detect a lesion, the output should be indistinguishable from the input. In the proposed model, an abnormal-to-normal generator (A2N-Generator) converts the input image to its healthy counterpart, and a discriminator is used to decide whether the input is a faked lesion-free image or a real healthy image.

We primarily test our model on the public Multimodal Brain Tumor Segmentation Challenge (BratS) dataset, which is based on magnetic resonance imaging (MRI) of the human brain. We also experiment on the Liver Tumor Segmentation Challenge (LiTS) dataset acquired by computational tomography (CT) on the human liver. Experiments on these two datasets consisting of different imaging modalities and human tissue demonstrate that ANT-GAN can produce highly realistic healthy-looking images corresponding closely to images containing lesions, which can aid the diagnostic work flow. We show some results on the BratS dataset in Figure \ref{figure1}: A real tumor MRI in Figure \ref{figure1a} is input into the well-trained A2N-Generator and the corresponding healthy-looking MRI is generated in Figure \ref{figure1b}. We take the absolute difference between the two images and give the color map in Figure \ref{figure1c} where we observe only the tumor regions are highlighted. We also input a real healthy MRI in Figure
\ref{figure1d} in the A2N-Generator and the corresponding output is shown in Figure \ref{figure1e}. The colormap of their absolute difference in Figure \ref{figure1f} shows little difference, indicating the generator doesn't detect a lesion that isn't there. Since we leverage a cycle consistency strategy in the ANT-GAN model, a normal-to-abnormal generator (N2A-Generator) can also be obtained. Such a regularization can not only stabilize the training and help convergence, but also provide a approach to encode the lesion information in the N2A-Generator, which has the potential in data augmentation where the training data is scarce.

\section{Related Work}
The generative adversarial network (GAN) is an emerging deep learning technique for modeling high-dimensional data distributions \cite{5} and has been widely used in computer vision tasks \cite{32,52,53,54}.
Image translation is one important application of GAN models \cite{15,16}. To overcome the need for perfectly aligned input-output pairs, CycleGAN \cite{17} uses a cycle consistency loss; an unsupervised approach is also taken by DualGAN \cite{18} and UNIT \cite{21} for image-to-image translation.

Medical image synthesis is becoming an active research topic in medical imaging. However, most existing work focuses on synthesizing across imaging modalities rather than restoring the image in some way. For example, \cite{22,24,25} map from MRI to CT, while \cite{23} map from MRI to PET and \cite{46} map from multiple MRI modalities to other modalities.
\cite{30} infer the manifold of normal tissue using a GAN architecture and develop an anomaly scoring scheme to predict abnormal tissue. In \cite{47} the GAN generates synthetic retinal images using segmentation labels, but pathological patterns are not considered.

More related to our paper, constrained adversarial auto-encoders are proposed in \cite{44} to detect lesions in brain MRI. The data distribution of brain MRI of healthy subjects are learned using a auto-encoder with unsupervised learning where the constraint that real lesion-containing medical data and its corresponding underlying lesion-free counterpart lie closely in latent space is also imposed. However, this approach is limited by the difficulty in handling high-resolution image synthesis as referred in this work.


\section{Methods}

\begin{figure}[t]
   {\includegraphics[width=1\columnwidth]{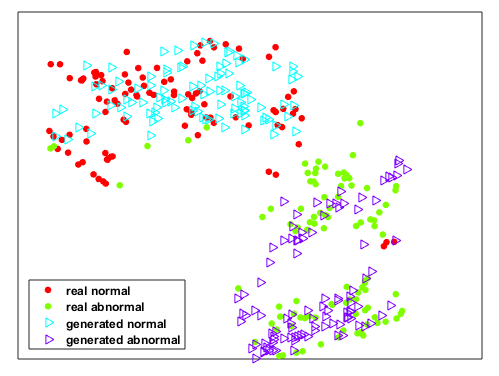}}
   \caption{100 normal (red) and 100 abnormal (green) medical images from the BratS dataset embedded in $\mathbb{R}^2$ using t-SNE. We also show their embeddings after mapping by their respective generators. Red maps to purple (N2A) and green maps to blue (A2N).}
\label{figure2}
\end{figure}

We denote a normal, healthy medical image as ${x^n}$ and an abnormal image with lesions as ${x^a}$. We assume the observed samples are drawn from their corresponding distributions, ${x^n} \sim {p_{n}}(x)$ and ${x^a} \sim {p_{a}}(x)$. In Figure \ref{figure2}, we show a t-SNE embedding of 100 true normal (red dot) and 100 true abnormal images (green dot) in the BratS dataset. To illustrate, we also show the learned A2N-Generator output of each abnormal image (blue triangle) and the learned N2A-Generator output of each normal image (purple triangle). We observe that the distance between the two normal and abnormal manifolds is small and we assume the difference is formed only by the lesions. We next present our ANT-GAN architecture that produced this result.

\subsection{ANT-GAN architecture overview}
\begin{figure*}[t]
\centering
   {\includegraphics[width=1\textwidth]{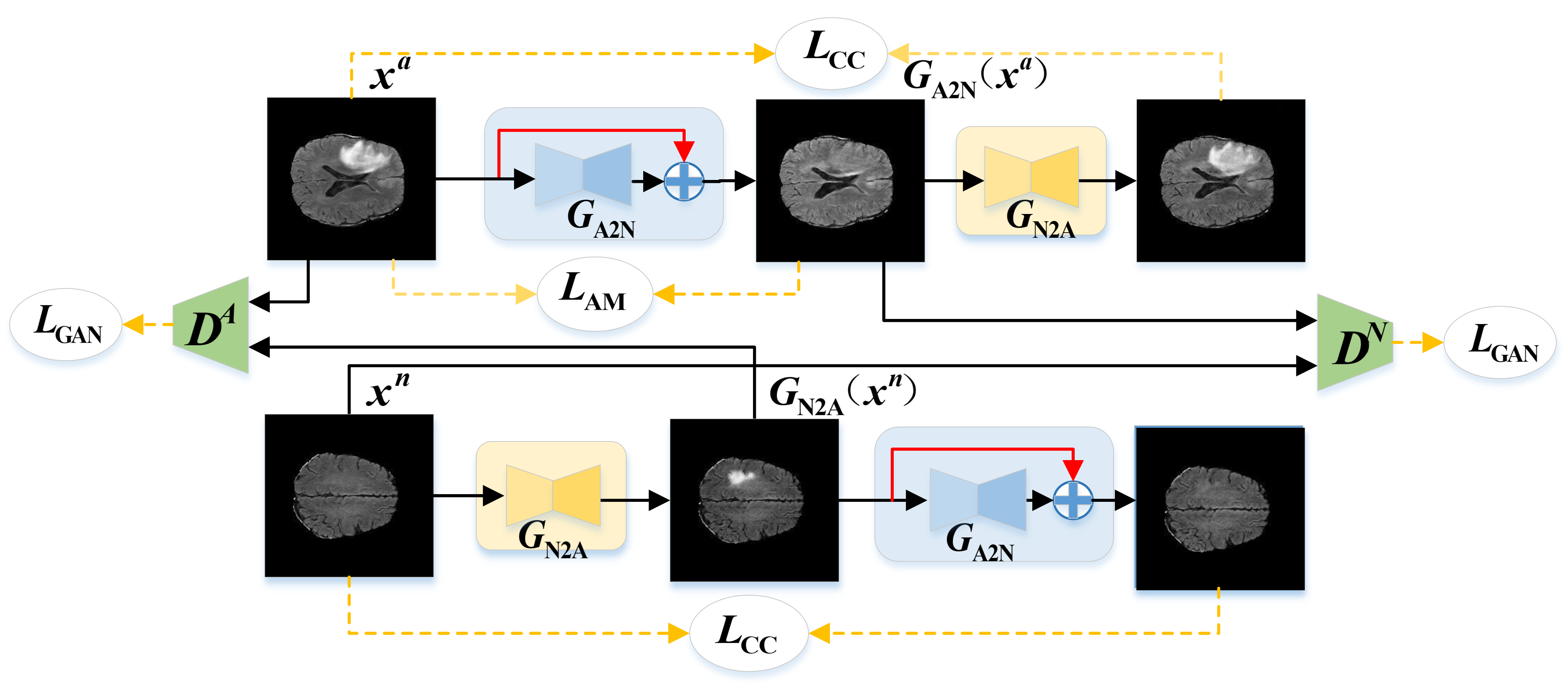}}
   \caption{The flowchart of our proposed ANT-GAN model. The data consists of measured abnormal and normal MRI or CT slices, $x^a$ and $x^n$ respectively. The other images represent intermediate steps within the model and are not measured data.}
\label{figure3}
\end{figure*}

The proposed ANT-GAN architecture can be described as an objective function consisting of three different parts, each motivated to capture one aspect desired by the problem. The flow chart of this architecture is shown in Figure \ref{figure3}. The motivation is to take a medical image as input and output a ``normal'' looking image corresponding to the input. If the input is healthy, we seek an output that is essentially unchanged from the input. The difference between the input and output can then be  used to segment abnormal regions and to classify healthy versus unhealthy images.

Our objective function consists of a standard GAN model, plus a cycle consistency loss and a problem-specific loss term to help isolate abnormal regions. The main deep network in our model is the generator $\mathcal{G}_{\rm{A2N}}$, which takes in an image $x$, assumed to contain an abnormal region but not necessarily so, and outputs the normal version $\mathcal{G}_{\rm{A2N}}(x)$. When ${\mathcal G}_{\rm{A2N}}$ is working well, it will produce realistic $\widehat x^n = {\mathcal G}_{\rm{A2N}}(x^a)$ to fool the discriminator ${\mathcal D}^N$. The generator $\mathcal{G}_{\rm{N2A}}$ and discriminator ${\mathcal D}^A$ are used to form the cycle consistency. However, different from many GAN implementations, not any realistic-looking $\widehat x^n$ is acceptable, but only one that looks like its corresponding $x^a$ with modifications in the abnormal regions. This motivates the following penalty, which we later validate via an ablation study.

Our full objective function consists of three terms and can be written as
\begin{equation}\label{eq4}
{\mathcal L_{\rm{FULL}}} = {\mathcal L_{\rm{GAN}}} + {\lambda _{\rm{CC}}}{\mathcal L_{\rm{CC}}} + {\lambda _{\rm{AM}}}{\mathcal L_{\rm{AM}}},
\end{equation}
which we minimize over $\mathcal{G}$ and maximize over $\mathcal{D}$. We break down each of these terms below.

\subsubsection{Term 1: Anomaly mask}
During training, we assume that each abnormal image has a corresponding binary mask provided with it that indicates where the abnormal locations are within the image. Let this mask be $\boldsymbol{\rm{M}}_x$, which is the same size as the image $x$ being considered in training. We emphasize here that this mask is not available and not needed during testing.

Since we want the generator $\mathcal{G}_{\rm{A2N}}$ to automatically isolate and modify the lesions within the image while leaving any healthy region within the image unchanged, we define the penalty
\begin{equation}\label{eq2}
{\mathcal L_{\rm{AM}}} = {\mathbb E_{{p_{a}}\left( x \right)}}\left[ {\left\| {\left( {\textbf{1} - \boldsymbol{\rm{M}}}_x \right) \odot \left( {{\mathcal G_{{\rm{A2N}}}}\left( {{x^a}} \right) - {x^a}} \right)} \right\|_2^2} \right],
\end{equation}
where $\odot$ represents element-wise multiplication and $\textbf{1}$ is an all-ones matrix of the same size of the input image. In other words, if the generator modifies a pixel or voxel in an abnormal image $x^a$ that does not correspond to the abnormal region, a heavy L2 penalty is paid.

\subsubsection{Term 2: GAN}
The main term in our objective is the GAN. Instead of building a unidirectional transform in the abnormal to normal direction, we adopt a bidirectional transform model with two generators ${\mathcal G_{{\rm{A2N}}}}$ and ${\mathcal G_{{\rm{N2A}}}}$ trained simultaneously. This strategy can help stabilize the model training via cycle consistency regularization. The trained ${\mathcal G_{{\rm{N2A}}}}$ can also produce highly realistic lesion-containing medical images, which has the potential for data augmentation not previously available for this problem. We have two generators and two discriminators, depending on whether the input data is normal $x^n$ or abnormal $x^a$.
\begin{equation}\label{eq3}
\begin{aligned}
\mathcal{L}_{\rm{GAN}} &= \mathbb{E}_{p_{a}}\left[\ln \mathcal{D}^A(x^a)\right] + \mathbb{E}_{p_{n}}\left[\ln \mathcal{D}^N(x^n)\right]\\
& + \mathbb{E}_{p_{n}}\left[\ln\, (1-\mathcal{D}^A(\mathcal{G}_{\rm{N2A}}(x^n)))\right]\\
& + \mathbb{E}_{p_{a}}\left[\ln\, (1-\mathcal{D}^N(\mathcal{G}_{\rm{A2N}}(x^a)))\right].\\
\end{aligned}
\end{equation}
We will discuss our selected networks for $\mathcal{D}$ and $\mathcal{G}$ in the following section. While ${\mathcal G_{{\rm{A2N}}}}(x^a)$ is trying to fool the discriminator ${\mathcal D}^N$, the ${\mathcal L_{\rm{AM}}}$ term teaches ${\mathcal G_{{\rm{A2N}}}}$ too fool it by only finding and modifying the abnormal regions. The adversarial training strategy is also adopted for ${\mathcal G_{{\rm{N2A}}}}$ and ${\mathcal D}^A$.

\subsubsection{Term 3: Cycle consistency}
As motivated by Figure \ref{figure2}, we adopt a cycle consistency term \cite{17} to transform normal and abnormal images into one another, and aid learning of ${\mathcal G_{{\rm{N2A}}}}$ and ${\mathcal G_{{\rm{A2N}}}}$,
\begin{equation}\label{eq1}
\begin{aligned}
{\mathcal L_{\rm{CC}}} &= {\mathbb E_{p_{a}}}\left[ {{{\left\| {\left( {{\mathcal G_{{\rm{N2A}}}}\left( {{\mathcal G_{{\rm{A2N}}}}\left( {{x^a}} \right)} \right) - {x^a}} \right)} \right\|}_1}} \right]\\
&+ {\mathbb E_{p_{n}}}\left[ {{{\left\| {\left( {{\mathcal G_{{\rm{A2N}}}}\left( {{\mathcal G_{{\rm{N2A}}}}\left( {{x^n}} \right)} \right) - {x^n}} \right)} \right\|}_1}} \right].
\end{aligned}
\end{equation}
This allows for additional information to be shared between normal and abnormal medical images when learning their corresponding generators. As part of this bidirectional regularization, we define the first term to be the abnormality synthesis consistency (AC), and the second to be normal synthesis (NC) consistency.
In the later ablation study, we show that bidirectional cycle consistency learns a better model than with either unidirectional consistency terms alone.

\subsection{Implementation}

\paragraph{Network Architecture and Training.}
For this medical imaging task in which accuracy is a major requirement of the model, the generator ${\mathcal G}_{\rm{A2N}}$ needs to detect and modify the lesion region while keeping other parts unchanged. The $\mathcal{L}_{\rm{AM}}$ penalty is meant to enforce this, but to further help in this task we include a global shortcut (the red arrow in Figure \ref{figure3}) to require the generator to learn a mapping that isolates and removes the lesion.

\begin{figure}[t]
\begin{center}
   \subfigure[\scriptsize Normal Brain]     {\label {figure4a} \includegraphics[width=0.1097\textwidth]{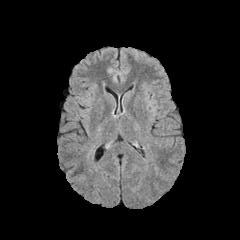}}
   \subfigure[\scriptsize Normal Brain]     {\label {figure4b} \includegraphics[width=0.1097\textwidth]{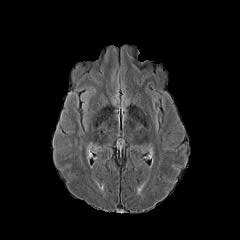}}
   \subfigure[\scriptsize Brain Tumor]      {\label {figure4c} \includegraphics[width=0.1097\textwidth]{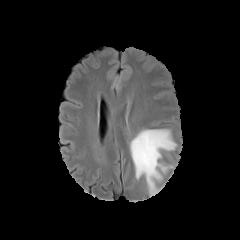}}
   \subfigure[\scriptsize Brain Tumor]      {\label {figure4d} \includegraphics[width=0.1097\textwidth]{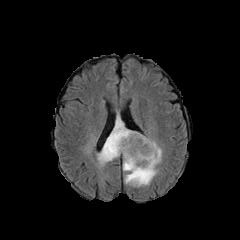}}
   \subfigure[\scriptsize Normal Liver]     {\label {figure5e} \includegraphics[width=0.1097\textwidth]{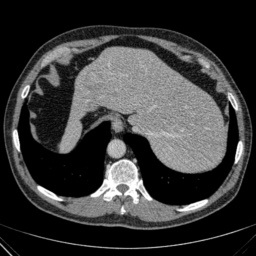}}
   \subfigure[\scriptsize Normal Liver]     {\label {figure5f} \includegraphics[width=0.1097\textwidth]{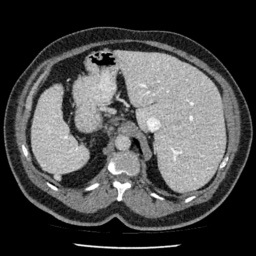}}
   \subfigure[\scriptsize Liver Tumor]      {\label {figure5g} \includegraphics[width=0.1097\textwidth]{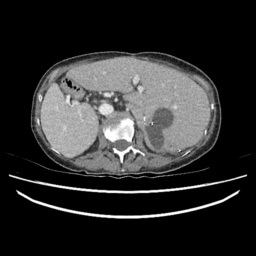}}
   \subfigure[\scriptsize Liver Tumor]      {\label {figure5h} \includegraphics[width=0.1097\textwidth]{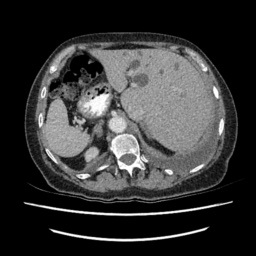}}
   \caption{We show some example brain MRI slices from BratS18 (first row) and Liver CT slices from LiTS (second row). Both normal- and abnormal-looking images are provided.}
\label {figure4}
\end{center}
\end{figure}

Following the proposal of \cite{17}, we also adopt the network architecture proposed in \cite{31} as the generators $\mathcal G_{\rm{N2A}}$ (except for the added global shortcut) and $\mathcal G_{\rm{A2N}}$. These two generators share the same network architecture but have different parameters. The generators consist of an encoder, residual blocks, and a decoder. We show the architecture of the generators in Table \ref{Gen}. In the generators, the 2-stride convolution in shallow layers under-samples the feature into smaller size and the 2-stride deconvolution up-samples the features to the input size. The residual blocks \cite{55} whose architecture is shown in Table \ref{RB} are also adopted to increase model capacity. Similarly, we use PatchGAN \cite{15,32} as the discriminators  $\mathcal D^{N}$ and $\mathcal D^{A}$. In this network, the classification problem is turned into a regression problem. The input image is mapped by the network to a $30 \times 30$ matrix, which is then compared against a matrix of all zeros or ones with an L2 penalty to stabilize training. The architecture of discriminators is shown in Table \ref{Dis}.  The instance normalization \cite{56} strategy is utilized to help accelerate training.

\begin{table}[t]
\center
\scriptsize
\caption{The network architecture of the generators. The ``FS" represents the filter size, ``IN" represents the instance normalization and ``Act" represents the activation function.}
\resizebox{\linewidth}{!}{
\begin{tabular}{|c|c|c|c|c|c|c|}
\hline
Layer     & Input       & FS & Stride & IN & Act & Output      \\ \hline
Conv1     & 240*240*1   & 7*7         & 1      & $\checkmark$  & ReLU       & 240*240*64  \\ \hline
Conv2     & 240*240*64  & 3*3         & 2      & $\checkmark$  & ReLU       & 120*120*128 \\ \hline
Conv3     & 120*120*128 & 3*3        & 2      & $\checkmark$  & ReLU       & 60*60*256   \\ \hline
RB$\times9$ & 60*60*256   &3*3         & 1      & $\checkmark$  & N/A           & 60*60*256   \\ \hline
Deconv1   & 60*60*256   & 3*3         & 2      & $\checkmark$  & ReLU       & 120*120*256 \\ \hline
Deconv2   & 120*120*256 & 3*3         & 2      & $\checkmark$  & ReLU       & 240*240*64  \\ \hline
Conv4     & 240*240*64  & 7*7         & 1      &               & tanh       & 240*240*1   \\ \hline
\end{tabular}}
\label{Gen}
\center
\end{table}

\begin{table}[h]
\center
\scriptsize
\caption{The architecture of the residual blocks (RB).}
\resizebox{\linewidth}{!}{
\begin{tabular}{|c|c|c|c|c|c|c|}
\hline
Layer & Input     & FS & Stride & IN & Act & Output    \\ \hline
Conv1 & 60*60*256 & 3*3         & 1      & $\checkmark$                       & ReLU        & 60*60*256 \\ \hline
Conv2 & 60*60*256 & 3*3         & 1      & $\checkmark$                       & N/A        & 60*60*256 \\ \hline
\end{tabular}}
\label{RB}
\center
\end{table}

\begin{table}[h]
\center
\scriptsize
\caption{The architecture of the discriminators.}
\resizebox{\linewidth}{!}{
\begin{tabular}{|c|c|c|c|c|c|c|}
\hline
Layer & Input      & FS  & Stride & IN & Act        & Output     \\ \hline
Conv1 & 240*240*1  & 4*4 & 2      &    & Leaky ReLU & 120*120*64 \\ \hline
Conv2 & 120*120*64 & 4*4 & 2      & $\checkmark$  & Leaky ReLU & 60*60*128  \\ \hline
Conv3 & 60*60*128  & 4*4 & 2      & $\checkmark$  & Leaky ReLU & 30*30*256  \\ \hline
Conv4 & 30*30*256  & 4*4 & 1      & $\checkmark$  & Leaky ReLU & 30*30*256  \\ \hline
Conv5 & 30*30*256  & 4*4 & 1      &    & N/A        & 30*30*1   \\ \hline
\end{tabular}}
\label{Dis}
\center
\end{table}

For training, we update the discriminator using the history of the previous $50$ generated images, rather than the output of the most recent generator. For our experiments, we set $\lambda_{\rm{CC}} = 10$, following \cite{17}. We also set $\lambda_{\rm{AM}} = 10$. To optimize, we use ADAM with batch size 1. The learning rate is set to $2\times 10^{-4}$ to train the networks for 20 epochs with 300K iterations.

\section{Results}
\subsection{Data}
We use two popular medical imaging datasets primarily used for the evaluation of lesion segmentation: the Multimodal Brain Tumor Segmentation Challenge 2018 dataset (BratS18) \cite{4,43} and the Liver Tumor Segmentation Challenge dataset (LiTS).

\paragraph{BratS18.}
The BratS18 dataset provides 210 high grade glioma (HGG) and 75 lower grade glioma (LGG) MRI with binary masks for the tumor (or lack of tumor). Each 3D MRI contains 155 slices of size $240 \times 240$. Not every slice contains a tumor, and therefore healthy MRI are provided by this data as well. We use the FLAIR modality image for all the experiments because the entire tumor is represented well by this modality.However, we also show more experimental results on other modalities, where the ANT-GAN provides impressive visual quality. A more detailed medical description of the data can be found on the challenge website.\footnote{\scriptsize \url{https://www.med.upenn.edu/sbia/brats2018.html}}

\begin{figure*}[t]
\begin{center}
   {\includegraphics[width=1\textwidth]{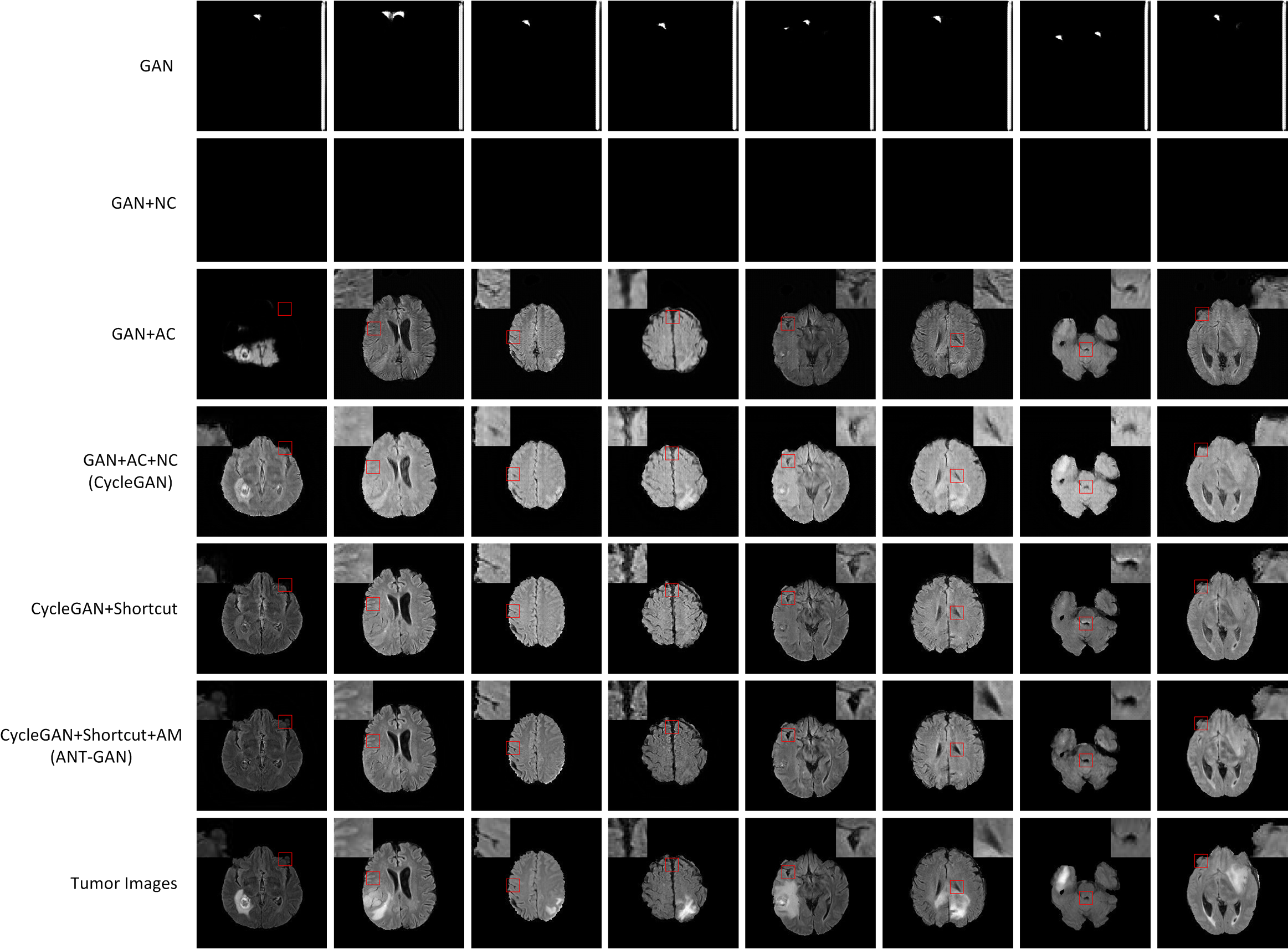}}
   \caption{The results of our ablation study on the BratS18 dataset. Please see text for analysis.}
\label {figure6}
\end{center}
\end{figure*}
\paragraph{LiTS.}
We also experiment with the LiTS data containing a total of 131 contrast enhanced abdominal CT volume images of the liver acquired from 7 different clinical institutions. The
in-plane resolution ranges from 0.5mm to 1mm and the slice thickness ranges from 0.7mm to 5.0mm. Each slice is $512 \times 512$ in size and we resize them to $256 \times 256$, and as with the BratS18 MRI not every slice contains a lesion and so these slices are considered to be healthy images. A detailed data description can be found on the challenge website. \footnote{\scriptsize \url{https://competitions.codalab.org/competitions/17094}}


Aside from the difference in imaging tissue and modality of these two data sets, the tumor regions on the CT images are of different shape and size, as can be seen in Figure \ref{figure4}. Also, many CT scans are acquired in a way that introduces greater noise-like artifacts than MRI. For each dataset, $80\%$ of randomly selected data are used for training and the resting $20\%$ for testing.

\subsection{Qualitative analysis}
Starting from the baseline GAN model, which consists of the $\mathcal{G}_{\rm{A2N}}$ generator without shortcut and AM loss and cycle consistency, we conduct the following ablation study to validate the cycle consistency, AM loss and shortcut.
\subsubsection{Evaluation of the cycle consistency}
We first conduct experiments to compare the baseline GAN \cite{5} to the same model, but with the abnormality or normality synthesis consistency penalty terms (GAN+AC and GAN+NC). In Figure \ref{figure6}, we show generated (i.e., fake) healthy-looking MRI produced by GAN, GAN+AC and GAN+NC. We observe that model collapse occurs in GAN and GAN+NC where the generator networks have converged to a bad local optimal solution. GAN+AC produces more meaningful image structures, however it still suffers severe artifacts due to the lack of optimizing constraints. We compare the above models with CycleGAN as a baseline state-of-the-art model for unsupervised image-to-image translation \cite{17}. The cycle consistency term reduces the artifacts by modifying the search space. However, the gray scale shift shows some bias.

\subsubsection{Evaluation of the shortcut and anomaly mask}
The global shortcut connection can simplify the function mapping by forcing the generator $\mathcal G_{\rm{A2N}}$ to focus on the lesion region. We compare with CycleGAN \cite{17} and CycleGAN with a shortcut connection (CycleGAN+shortcut). The difference between CycleGAN+shortcut and our ANT-GAN is the inclusion of the anomaly mask penalty term.

In Figure \ref{figure6}, we observe that the generator can better detect, remove and inpaint the tumor regions without impacting the non-tumor regions by using the proposed global skip connection. However, as shown by the zoomed-in regions, CycleGAN+shortcut still performs less satisfactorily than ANT-GAN in terms of some important details of the healthy regions of the lesion-containing MRI. This is because ANT-GAN contains the anomaly mask term, which forces the generator $\mathcal{G}_{\rm{A2N}}$ to leave healthy portions of an MRI unchanged and only detect and modify lesions. Both CycleGAN and CycleGAN+shortcut do not have this feature since they have different motivation in their design. Though based on the GAN, ANT-GAN is more motivated by image restoration than image generation.

\begin{figure}[t]
\begin{center}
   \subfigure[ Conv1 Feature] {\includegraphics[width=0.49\columnwidth]{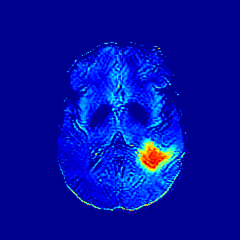}}
   \subfigure[ Conv1 Feature] {\includegraphics[width=0.49\columnwidth]{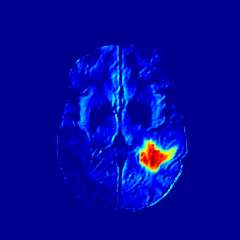}}
   \subfigure[ Deconv2 Feature] {\includegraphics[width=0.49\columnwidth]{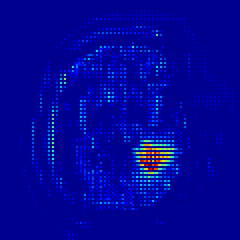}}
   \subfigure[ Deconv2 Feature] {\includegraphics[width=0.49\columnwidth]{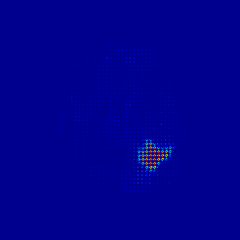}}
   \caption{A visualization of two features of the generator $\mathcal{G}_{\rm{A2N}}$ on the BratS MRI data. The tumor is gradually identified.}
\label {figure111}
\end{center}
\end{figure}

We show feature maps of the generator $\mathcal{G}_{\rm{A2N}}$ on a BratS MRI data in Figure \ref{figure111}, which clearly shows that the generator $\mathcal{G}_{\rm{A2N}}$ tries to capture and work on the lesion regions with the global shortcut connection.

We also use PSNR metric to further validate the benefit of shortcut and AM loss strategies in preventing non-lesion regions from distortion. The results evaluated on the testing datasests in BratS18 are shown in Table \ref{PSNR}. We observe the proposed ANT-GAN (ANT + AC + NC + Shortcut + AM) achieves the highest PSNR value in the compared models, proving the effectiveness of the shortcut architecture and AM loss in non-lesion region preservation. The lesion regions are not amenable to objective evaluation since their ground truth healthy counterpart are unknown naturally.

\begin{table}[t]
\caption{The objective assessment on the shortcut and AM loss strategy using PSNR metric.}
\resizebox{\linewidth}{!}{
\begin{tabular}{|l|c|}
\hline
\multicolumn{1}{|c|}{Model} & PSNR dB \\ \hline
GAN + AC + NC (CycleGAN)        & 21.34   \\ \hline
GAN + AC + NC + Shortcut         & 27.29   \\ \hline
GAN + AC + NC + Shortcut + AM     & 28.44   \\ \hline
\end{tabular}}
\label{PSNR}
\end{table}

\subsubsection{Comparison with other GAN formulations}
\begin{figure}[t]
   {\includegraphics[width=1\columnwidth]{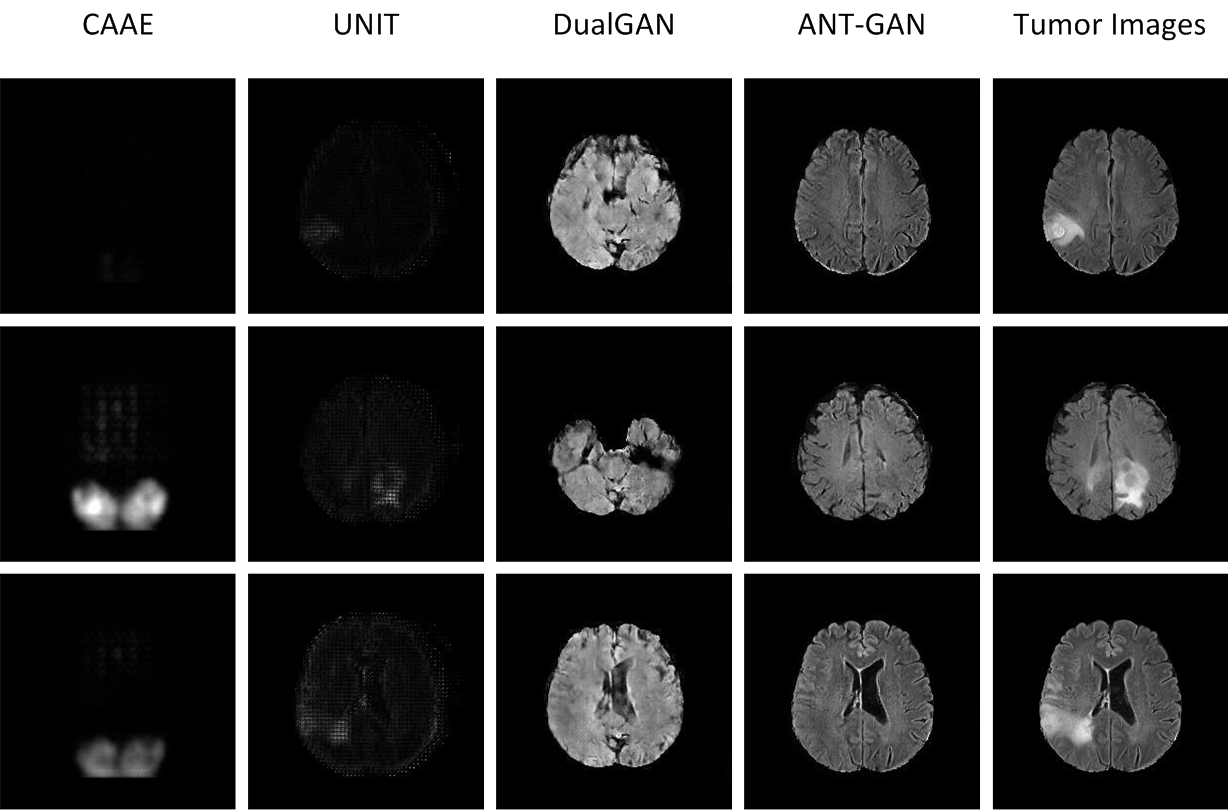}}
   \caption{We compare ANT-GAN with other state-of-the-art GAN methods less tailored to this problem.}
\label {figure7}
\end{figure}

We compare the proposed ANT-GAN model with the prior work using constrained adversarial auto-encoder model (CAAE) for lesion detection \cite{44} and other two recently proposed state-of-the-art unsupervised GAN models, UNIT \cite{20} and DualGAN \cite{18}. We show these results in Figure \ref{figure7}. We observe that UNIT does not work for this problem. The data sets in this case are too small and the images too large for these models to learn in their less-regularized settings. ANT-GAN also outperforms DualGAN in imaging quality. While DualGAN uses cycle consistency, which makes learning $\mathcal{G}$ easier with less data, no shortcut in $\mathcal{G}$ is used by DualGAN, unlike ANT-GAN. Finally, the stricter regularization of the anomaly mask in ANT-GAN (absent from all other GAN models) not only can enforce greater fidelity to the original image, which is required for this problem, but also aid GAN learning by introducing greater supervision. We observe that CAAE struggles to produce high
quality normal-looking medical images in such a high-resolution image synthesis task, which is also the main limitation mentioned in that paper, where evaluations are performed on much smaller images of size $32\times32$.

\subsection{Practical Applications}
In this section, we experiment with using the output of ANT-GAN as input to image segmentation and classification models. We compare with the performance of these same models using the original medical image only without additional information provided by ANT-GAN. The purpose of these experiments is to show how ANT-GAN can supplement existing models to improve their results.

\paragraph{Application to image segmentation.}
The expert radiologist makes diagnoses based on prior knowledge about characteristics of healthy and unhealthy images. To assist in this analysis, automatic segmentation has become an important task in the field of medical image analysis \cite{34}. Since ANT-GAN is trained to isolate abnormal tumor regions and fix those regions only, the difference between an input image $x$ and output image $\mathcal{G}_{\rm{A2N}}(x)$ can be used to segment MRI for areas of potential concern.

\begin{figure}[t]
   {\includegraphics[width=1\columnwidth]{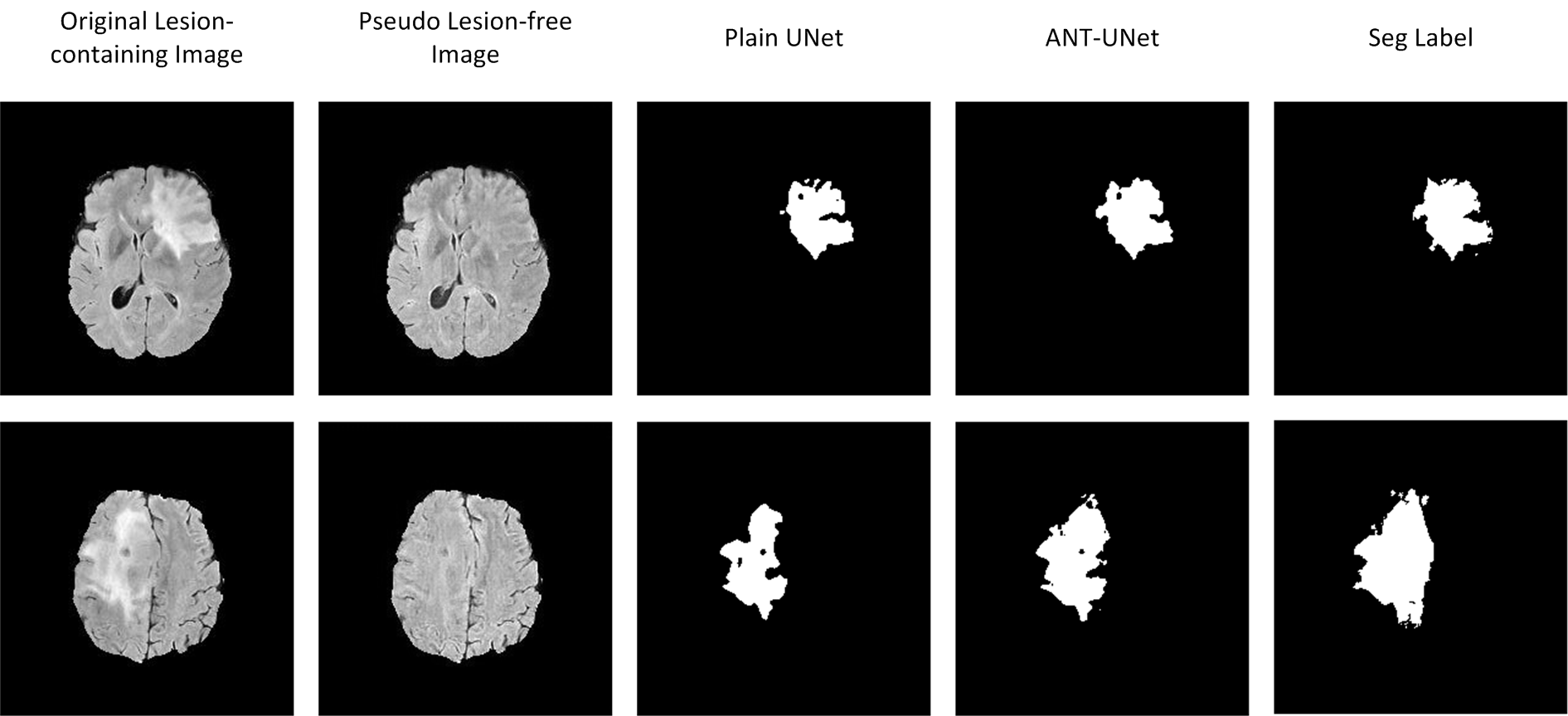}}
   \caption{Example segmentation of two MRI slices.}
\label {figure8}
\end{figure}

For the segmentation model, we use the state-of-art UNet \cite{3} to segment each slice. We input to UNet each MRI slice $x$ and its corresponding generated lesion-free MRI slice $\mathcal{G}_{\rm{A2N}}(x)$ from an already-trained ANT-GAN model as multi-channel inputs (referred to as ANT-UNet). We compare with UNet in which only $x$ is input without using information from $\mathcal{G}_{\rm{A2N}}(x)$ (referred to as Plain UNet). To ensure that the number of parameters is the same in both models for fair comparison, we use a copy of the MRI slice $x$ to create a multi-channel input for Plain UNet.
We train the ANT-GAN model first, and then ANT-UNet and Plain UNet on BratS18 using 5-fold cross validation. In Table \ref{Table0} we compare their segmentation performances using the Dice Coefficient (larger is better). The improvement demonstrates that the prior information provided by ANT-GAN on where the lesion may be can significantly aid predictions made by the state-of-the-art segmentation model UNet. We show two visual examples in Figure \ref{figure8}.

\begin{table}[h]
\caption{The performance comparison between the Plain UNet and ANT-UNet on BratS18.}
\resizebox{\linewidth}{!}{
\begin{tabular}{|l|c|c|c|}
\hline
DC & Enhancing Tumor & Whole Tumor & Tumor Core  \\ \hline
Plain UNet       & 75.44\%         & 87.60\%    & 77.53\%     \\ \hline
ANT-UNet         & 77.77\%           & 89.10\%    & 79.77\%     \\ \hline
\end{tabular}
}
\label{Table0}
\end{table}

\begin{figure}[t]
   \subfigure[ Tumor $x$] {\includegraphics[width=0.113\textwidth]{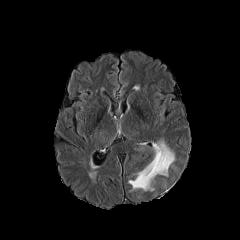}}
   \subfigure[ $\mathcal{G}_{\rm{A2N}}(x)$] {\includegraphics[width=0.113\textwidth]{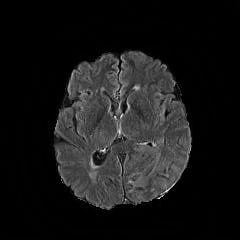}}
   \subfigure[ Prediction] {\includegraphics[width=0.113\textwidth]{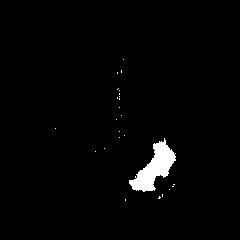}}
   \subfigure[ Label] {\includegraphics[width=0.113\textwidth]{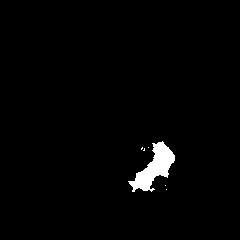}}
   \caption{Example segmentation obtained by taking the absolute difference between the real tumor MRI $x$ and the generated normal-looking MRI $\mathcal{G}_{\rm{A2N}}(x)$, after binarization at a preset threshold.}
\label {figure88}
\end{figure}

We observe that the generated normal-looking from ANT-GAN can also be used to directly segment the image, since the only difference between a synthesized normal-looking image and its real abnormal counterpart is region with the lesion. To illustrate this, we calculate the absolute difference between $x$ and $\mathcal{G}_{\rm{A2N}}(x)$ and show the segmentation after binary thresholding at 0.1 in Figure \ref{figure88}.

\paragraph{Application to image classification.}
We also evaluate the benefits of using the output of ANT-GAN for a lesion classification task. As classifier, we adopt the deep model VGG \cite{40} as the base classifier to predict if the input image contains a lesion or not. Again, to ensure the same number of parameters for comparison, we use a duplicate of $x$ in place of $\mathcal{G}_{\rm{A2N}}(x)$ for Plain VGG, while we input both $x$ and $\mathcal{G}_{\rm{A2N}}(x)$ as input to the VGG (referred to as ANT-VGG) to see if $\mathcal{G}_{\rm{A2N}}(x)$ brings any additional discriminative information. We show the classification results in the Table \ref{Table1}. We use a 5-fold cross validation for evaluation of BratS18. We observe that ANT-VGG outperforms the Plain VGG in all three classification metrics, showing that ANT-GAN can improve the medical image classification task for detecting lesions.

\begin{table}[h]
\caption{Comparison for lesion classification.}
\resizebox{\linewidth}{!}{
\begin{tabular}{|l|c|c|c|}
\hline
Methods   & Precision       & Recall      & F1-Measure \\ \hline
Plain-VGG &    89.41\%       &   89.86\%     & 0.896           \\ \hline
ANT-VGG   &    92.35\%       &   90.96\%     & 0.917           \\ \hline
\end{tabular}
}
\label {Table1}
\end{table}


\subsection{Results on more MRI modalities with brain tumor.}
\begin{figure*}[!htbp]
\begin{center}
   {\includegraphics[width=0.88\textwidth]{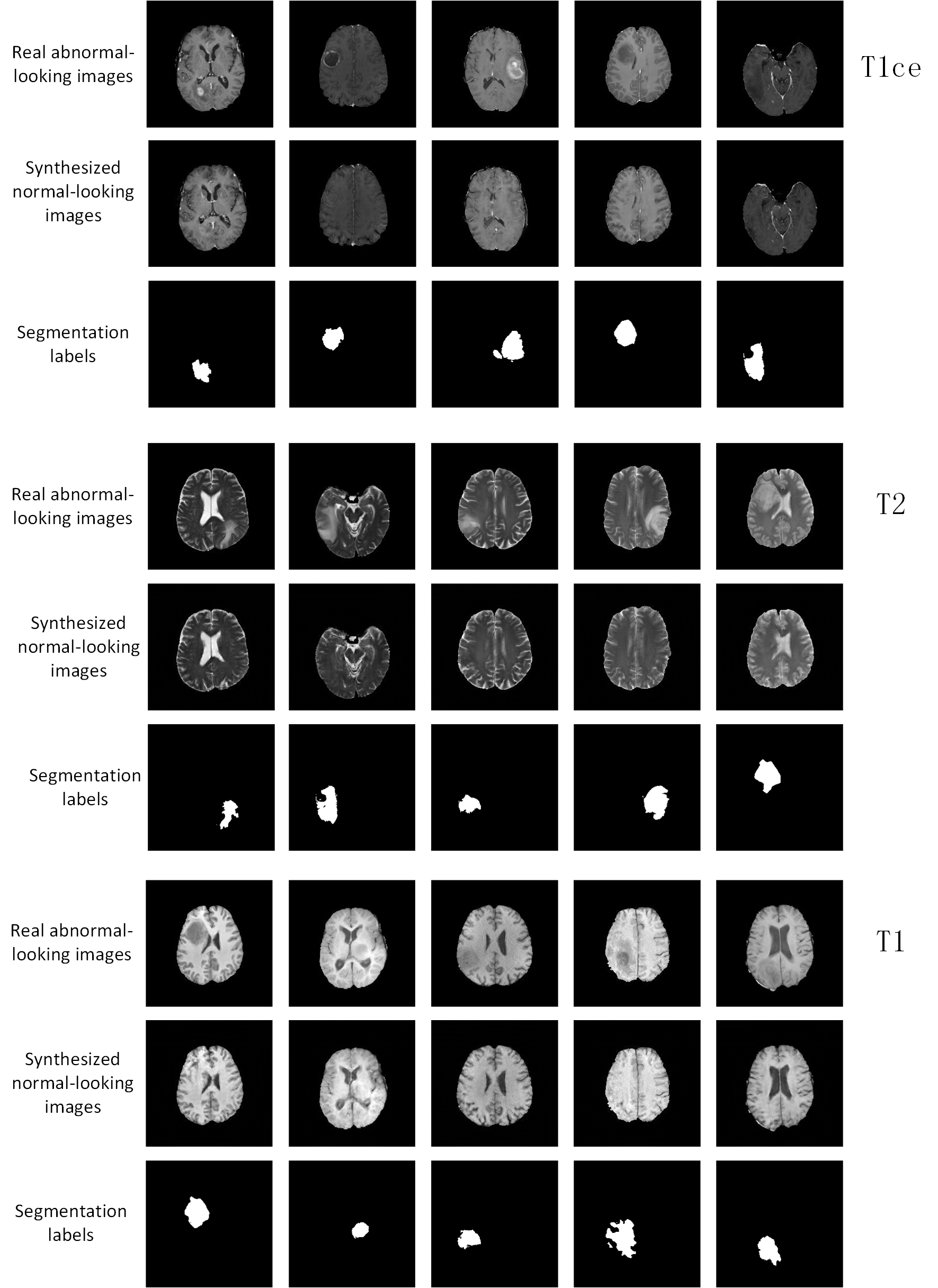}}
   \caption{More results on the synthesized normal-looking MRI images on T1ce, T2 and T1 modalities.}
\label {OM}
\end{center}
\end{figure*}

The ANT-GAN model is mainly evaluated and validated on the FLAIR modality. However, we also test the ANT-GAN model on other three MRI setting including T2, T1ce and T1 modalities. The generated pseudo healthy images are shown in Figure \ref{OM}.

%
%

\subsection{Results on the LiTS Challenge dataset.}
\begin{figure}[h]
   {\includegraphics[width=1\columnwidth]{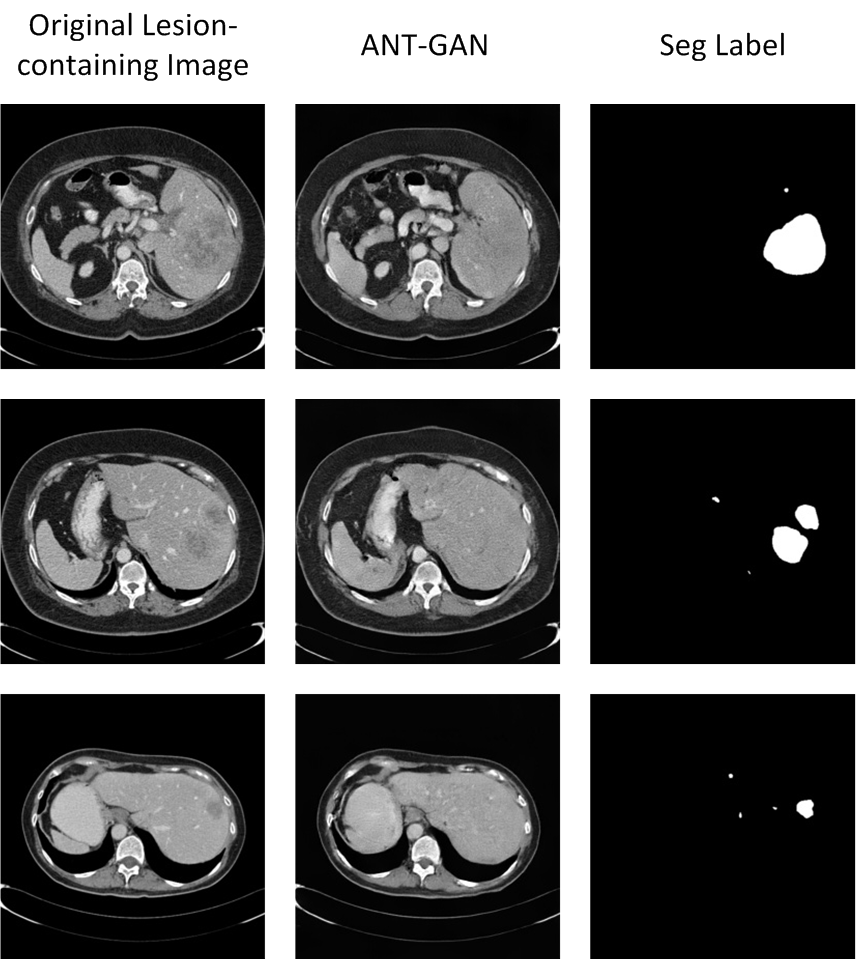}}
   \caption{Experimental results on the Liver CT data (LiTS).}
\label{figure9}
\end{figure}

We also implement ANT-GAN on the LiTS dataset and show some qualitative results in Figure \ref{figure9}. We observe that the lesions in the liver CT data appears with much lower contrast than in the brain MRI data. While our model can detect and modify the abnormal regions successfully, we note that there are more deformations than with the BratS18 dataset, which is a result of this more difficult task.

\section{Discussions}
\subsection{Sensitivity to anomaly mask parameter $\lambda_{\rm{AM}}$}
\begin{figure}[t]
       \subfigure[ Tumor] {\label {figure11a} \includegraphics[width=0.312\columnwidth]{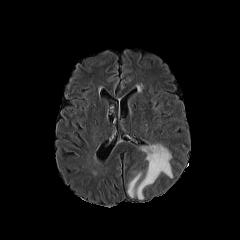}}
       \subfigure[ $0.01$] {\label {figure11b} \includegraphics[width=0.312\columnwidth]{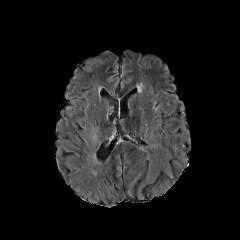}}
       \subfigure[ $0.1$] {\label {figure11c} \includegraphics[width=0.312\columnwidth]{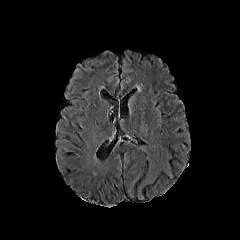}}
       \subfigure[ $1$] {\label {figure11d} \includegraphics[width=0.312\columnwidth]{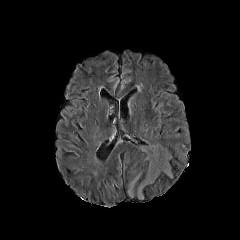}}
       \subfigure[ $10$] {\label {figure11e} \includegraphics[width=0.312\columnwidth]{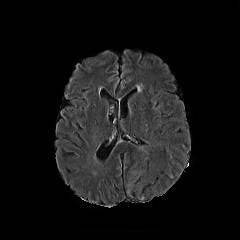}}
       \subfigure[ $100$] {\label {figure11f} \includegraphics[width=0.312\columnwidth]{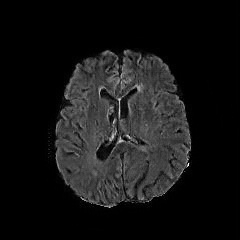}}
    \caption{The results produced with different $\lambda_{\rm{AM}}$.}
\label{figure11}
\end{figure}

We discuss how the regularization parameter $\lambda_{\rm{AM}}$ for the anomaly mask term influences the result of our ANT-GAN model and show these results in Figure \ref{figure11}. We observe that setting $\lambda_{\rm{AM}}=10$ leads to a good balance between the preservation of non-lesion regions and the modification of the lesion.

\subsection{Synthesizing abnormal-looking images}
The ability of generating highly realistic pseudo healthy MR image ANT-GAN model using the generator $\mathcal{G}_{\rm{A2N}}$ has been verified in our experiments. However, the well-trained ANT-GAN model can produce another generator $\mathcal{G}_{\rm{N2A}}$.

\begin{figure}[t]
       \subfigure[ Truth] {\label {figure12a} \includegraphics[width=0.48\columnwidth]{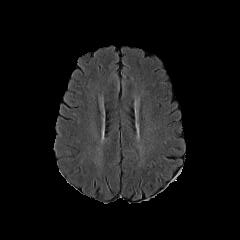}}
       \subfigure[ Synthesized lesion] {\label {figure12b} \includegraphics[width=0.48\columnwidth]{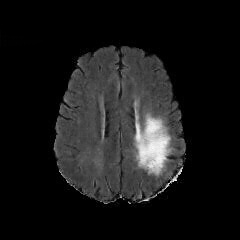}}
       \subfigure[ Truth] {\label {figure12c} \includegraphics[width=0.48\columnwidth]{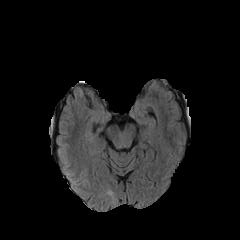}}
       \subfigure[ Synthesized lesion] {\label {figure12d} \includegraphics[width=0.48\columnwidth]{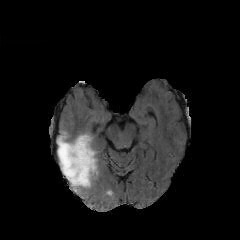}}
    \caption{Synthesis of a lesion $\mathcal{G}_{\rm{N2A}}(x)$ in two healthy MRI $x$.}
\label{figure12}
\end{figure}

We leverage the trained $\mathcal{G}_{\rm{N2A}}$ to synthesize abnormal-looking images on BratS18 datasets. Two examples are shown in Figure \ref{figure12}. We observe that the tumor is synthesized near the brain boundary in the second example. In training $\mathcal{G}_{\rm{N2A}}$, the discriminator $\mathcal{D}^{\rm{A}}$ encodes the core patterns of the real abnormal-looking images data, which can avoid having lesions being generated in physiologically unreasonable locations. We observe the generator $\mathcal{G}_{\rm{N2A}}$ produces realistic lesion-containing MR images.

The ANT-GAN model is capable of synthesizing pseudo normal and abnormal images, showing a potential in augmenting data in some cases where the medical images are scarce. To evaluate such potential, we randomly draw 50 real lesion-containing MR images and 50 real lesion-free ones from BratS18 datasets to form a small-scale datasets A. We utilize the chosen 50 real lesion-free and lesion-containing MR images in datasets A to predict their abnormal and normal counterparts using trained $\mathcal{G}_{\rm{N2A}}$ and $\mathcal{G}_{\rm{A2N}}$, yielding 50 faked healthy images and 50 faked unhealthy ones. We generate a datasets B with the data contained in datasests A plus the pseudo 50 abnormal and 50 normal images.

We train two AlexNet \cite{57} models to identify if the image contains any lesion with training on Datasets A and Datasets B, and the two trained AlexNet models are tested on the resting data in the BratS18 datasets. We show the averaged results on the classification metrics in Table \ref{Alex}.

\begin{table}[]
\caption{The evaluation of the AlexNet model trained on 2 datasets for lesion identification.}
\resizebox{\linewidth}{!}{
\begin{tabular}{|l|c|c|c|}
\hline
Training Datasets           & Precision & Recall  & F1-Measure \\ \hline
Datasets A with 100 samples & 65.56\%   & 63.36\% & 0.644      \\ \hline
Datasets B with 200 samples & 77.19\%   & 77.94\% & 0.776      \\ \hline
\end{tabular}}
\label{Alex}
\end{table}

\section{Conclusion}
We proposed an generative adversarial network called ANT-GAN for translating a medical image containing lesions into a corresponding image where the lesion has been ``removed'' via color correction. We showed how being able to generate these two versions of the same image can help in the medical image segmentation and classification tasks, since the generator can provide additional information about what the image ``should'' look like if it were healthy. We also showed how our generator was able to not be fooled by healthy MRI, in which case it simply output a near replication of the input image when no lesion is present.

Our objective function was tailored to the problem by introducing an binary anomaly mask term that indicates the lesion location, as well as cycle consistency constraint to regularize the space. (We again note that this mask is not needed for test images.) Comparison with other GAN setups showed how this was a requirement to successfully learn from this small data set. Experiments on the BratS18 and LiTS challenge data sets helped to validate our framework for using computer vision to address fundamental medical image analysis problems of segmentation and classification.

\small
\bibliographystyle{ieee}
\bibliography{ANTGAN}

\end{document}